\documentclass[letterpaper, 10 pt, conference]{ieeeconf}  

\IEEEoverridecommandlockouts

\usepackage{cite}
\usepackage[pdftex]{graphicx}
\usepackage{amsmath}
\usepackage{algorithmic}
\usepackage{url}
\usepackage{multirow}
\usepackage{subfigure}
\usepackage{amsfonts}

\usepackage{mathtools}
\usepackage{bm}

\newcommand\Set[2]{\{\,#1\mid#2\,\}}
\newcommand\SET[2]{\Set{#1}{\text{#2}}}

\usepackage[colorinlistoftodos]{todonotes}

\newif\ifcommentsandstuff

\commentsandstufftrue

\usepackage{soul}

\title{\LARGE \bf Multi-camera Torso Pose Estimation using Graph Neural Networks}

\author{Daniel Rodriguez-Criado$^{1}$, Pilar Bachiller$^{2}$, Pablo Bustos$^{2}$, George Vogiatzis$^{1}$, and Luis J. Manso$^{1}$%
\thanks{$^{1}$Daniel Rodriguez-Criado, George Vogiatzis and Luis J. Manso are with School of Engineering and Applied Science, Computer Science Department, Aston University, United Kingdom.
{\tt\footnotesize l.manso@aston.ac.uk}}%
\thanks{$^{2}$Pilar Bachiller and Pablo Bustos are with Robotics and Artificial Vision Laboratory, Caceres School of Technology, Universidad de Extremadura, Extremadura.}%
}

\begin{document}

\maketitle
\thispagestyle{empty}
\pagestyle{empty}

\begin{abstract}
Estimating the location and orientation of humans is an essential skill for service and assistive robots.
To achieve a reliable estimation in a wide area such as an apartment, multiple RGBD cameras are frequently used.
Firstly, these setups are relatively expensive.
Secondly, they seldom perform an effective data fusion using the multiple camera sources at an early stage of the processing pipeline.
Occlusions and partial views make this second point very relevant in these scenarios.
The proposal presented in this paper makes use of graph neural networks to merge the information acquired from multiple camera sources, achieving a mean absolute error below $\bm{125 mm}$ for the location and $\bm{10^\circ}$ for the orientation using low-resolution RGB images.
The experiments, conducted in an apartment with three cameras, benchmarked two different graph neural network implementations and a third architecture based on fully connected layers.
The software used has been released as open-source in a public repository\footnote{https://github.com/vangiel/WheresTheFellow}.

\end{abstract}
\section{INTRODUCTION}\label{intro}
Autonomous robots have a wide range of applications, including performing daily chores for an ageing population and carrying out tasks that might be dangerous for humans.
To work seamlessly among humans, robots need social skills, for instance, to not to get in the way, or to understand people's intentions and communicate their own.
Among other relevant information such as gestures or facial expressions, people's position and orientation are among the most important cues that can help service and assistive social robots understand humans.
A common application of human localisation and orientation is predicting intentions and movements in surveillance video feeds~\cite{Glonek2017,Thang2019}.
An accurate localisation and orientation estimation are also crucial for human-aware navigation~\cite{manso2019graph}.
For instance, the orientation of pedestrians' velocity vectors is used in~\cite{Mateus2019} to make a robot navigate in crowded environments complying with constraints defined by proxemics.
\par

Although there is a considerable number of exceptions (\textit{e.g.},~\cite{Shinmura2015, Wahyu2019, Lewandowski2020}) orientation and other social cues are usually acquired using a two-stage pipeline: human body parts are detected as a first step and then passed as input to a second stage algorithm.
This second algorithm is frequently implemented using basic trigonometry, considering the coordinates of the shoulders or the hips~\cite{Glonek2017}.
For instance, in~\cite{Choi2016}, it is calculated using the cross-product of the vectors going from the head to the right and left sides of the hip, respectively.
\par

To overcome the poor behaviour that handcrafted equations tend to have when working with missing, noisy and redundant data, some works follow a machine \mbox{learning-based} approach.
For instance, \cite{Shinmura2015},  and~\cite{Wahyu2019} use Histograms of Oriented Gradients (HOGs).
In~\cite{Shinmura2015} RGBD HOGs were used to provide discrete angle estimations.
The work presented in~\cite{Lewandowski2020} uses RGBD and IR images with IR trackers to train a single-camera Convolutional Neural Network (CNN).
Their model provides continuous angle estimation, achieving a mean absolute error close to $6^{\circ}$.
The accuracy of the work at hand for torso orientation is below that of~\cite{Lewandowski2020}.
However, this proposal does not only estimate the orientation but also the 3D coordinates of the torsos and it does not require the use of relatively expensive RGBD cameras.
To do this, our work builds on top of a skeleton detector and Graph Neural Networks (GNNs).
To the best of our knowledge, there are no previous GNN models to predict pedestrian's orientation.
\par

There are numerous works on human detection.
Pioneering works such as~\cite{Pfinder} or~\cite{Papageorgiou2000} took RGB images as input.
With the advent of RGBD cameras, different alternatives were made available in the early 2010s~\cite{shotton2011real,liu2015detecting}.
The additional depth channel made these algorithms less sensitive to illumination changes and made some tasks such as segmentation more approachable. 
Nevertheless, they had important limitations when applied to robotics such as a low accuracy distinguishing the left and right sides of humans in specific angle ranges and poor performance on moving cameras~\cite{calderita2013model}.
Many works have been recently published using CNNs to address some of the limitations of previous approaches.
OpenPose~\cite{cao2018openpose} became state-of-the-art detecting body parts using Part Affinity Fields to learn the association between body parts and humans in the image.
However, its performance deteriorates as resolution decreases and does not work as well in crowded environments with occluded body parts.
OpenPifPaf~\cite{kreiss2019pifpaf} was proposed to solve OpenPose's limitations. 
It uses a CNN with two heads; the first to locate the joints and the second to predict associations between them, including occluded parts.


\section{MULTI-CAMERA TORSO POSE ESTIMATION} \label{problem}
\begin{figure*}[!ht]
  \centering
  \subfigure{
  \includegraphics[width=.28\textwidth]{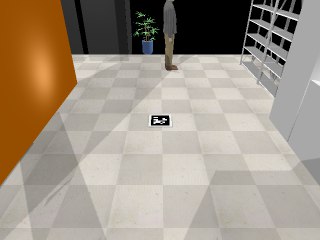}
  \label{fig:view_cam1}
  }
  \subfigure{
  \includegraphics[width=.28\textwidth]{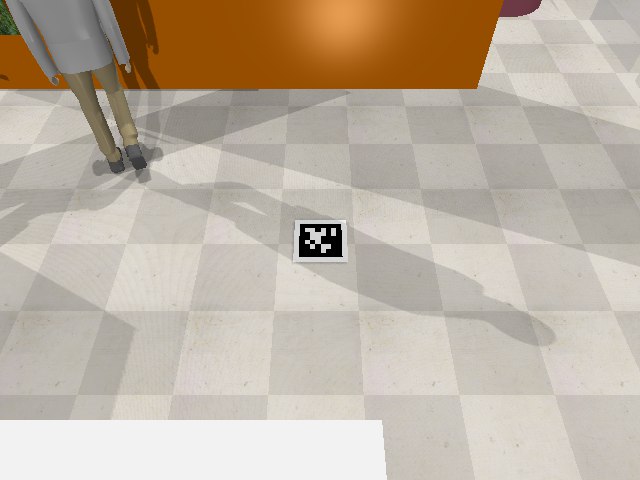}
  \label{fig:view_cam2}
  }
  \subfigure{
  \includegraphics[width=.28\textwidth]{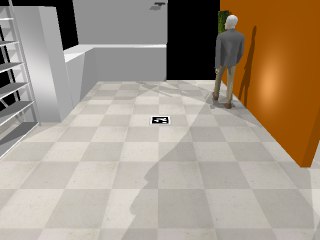}
  \label{fig:view_cam3}
  }
  \caption{Example images obtained from the simulator. In the experiments, the resolution used was 640x480.}
  \label{fig:camera_views}  
\end{figure*}

The problem we deal with is that of estimating the pose of a person from a set of cameras.
The pose is defined as their position on the floor plane and their orientation with respect to the vertical axis: $\left(x, y, \alpha\right)$.
The system should be able to cover spaces wide enough to require several cameras attached to the walls with overlapping fields of view.
The setup used in the experiments is composed of 3 Intel RealSense 415 depth cameras whose extrinsic parameters are calibrated with respect to a common reference system on the floor (RGBD cameras are used to allow comparing results using RGB and RGBD images).
This paper uses both real and synthetic training data to generate a large dataset in a very short time, saving a great amount of resources.
Figure~\ref{fig:camera_views} shows a representation in CoppeliaSim~\cite{rohmer2013coppeliasim} of three views of the environment where the system has been tested.
As shown below, once the model is trained it can estimate 3D poses using only the joints' image coordinates, not requiring depth data -although it can optionally be used to enhance results.

\par
The processing pipeline has three main stages.
First, images are acquired and processed using OpenPifPaf~\cite{kreiss2019pifpaf}.
The output data of this stage -a set of detected skeletons from the different cameras- is passed to the next stage, where skeletons corresponding to the same person are matched and grouped.
These groups are then provided to a GNN which provides the final output.
The remainder of this section explains the stages in more detail.
\par

\textbf{Image acquisition and skeleton detection}: The images acquired are provided to OpenPifPaf to get the skeleton data.
For each frame, an observation $\Psi \left\{ p_i, r_i, t_i\right\}$ is generated and provided to the next stage, where:
\begin{itemize}
    \item $p_i$ is the set of people detected in that frame, each of which holds a list of up to 16 joint's coordinates. If using RGBD cameras, each joint's depth from the camera is computed using the depth plane.
    \item $r_i$ is the RGB Region of Interest (ROI) corresponding to the bounding box of the skeleton.
    \item $t_i$ is the acquisition time of the frame.
\end{itemize}
\par

\textbf{Match observations to people:} a stream of $\Psi_t, \; t \in \mathbb{N}$, observations are generated from the skeleton detectors.
A state machine manages the creation, update and removal of a set of data objects representing people. Each observation can either create a new person, update an existing one or be dismissed as noise.
Before a new person is accepted, it has to receive successful matches for at least 2 seconds.
An observation matches an existing person if their distance $d(o_i, p_j)$ is lower than a certain threshold $d_{max}$, taken here as $0.65$ in the $\left[0,1\right]$ range.
The distance is defined as the median of the distances between the observation and the recent history of the person: 
$ d = med_{t}\left\{ B\left(h(o_i), h_t(p_j) \right)\right\}$
where $B$ is the Bhattacharyya distance~\cite{kailath1967divergence}, $h(o_i)$ is the observation's 2D histogram computed over the hue and saturation planes of the person's ROI and $h_t(p_j)$ is the 2D histogram of person $p_j$ at time $t$, where $t$ goes from the last observation to $Q$ samples in the past. 
Other distances have been tested with no better results. The removal of unseen people occurs after 2 seconds without receiving any matches.
\par
In the next stage, a set $S$ of observations from a person  is fed to the GNN to obtain a tuple of target coordinates $\left( x,y,\alpha \right)$ representing the pose for each torso. This set $S$ is extracted from the person's history as:
\begin{align*}
    S&\coloneqq \SET{o}{$o_i^k \in O^p \; \wedge k \in \{1,2,3\} $}
\end{align*}
where $o_i^k$ is the past matched observation $i$ by camera $k$ and $O^p$ is the set of past time-ordered observations of person $p$. $S$ is thus the set of the most recent matched observations obtained from different cameras. 
\par

\begin{figure}[!ht]
  \centering
  \subfigure[Person detected by the three cameras.]{
  \includegraphics[width=.75\columnwidth]{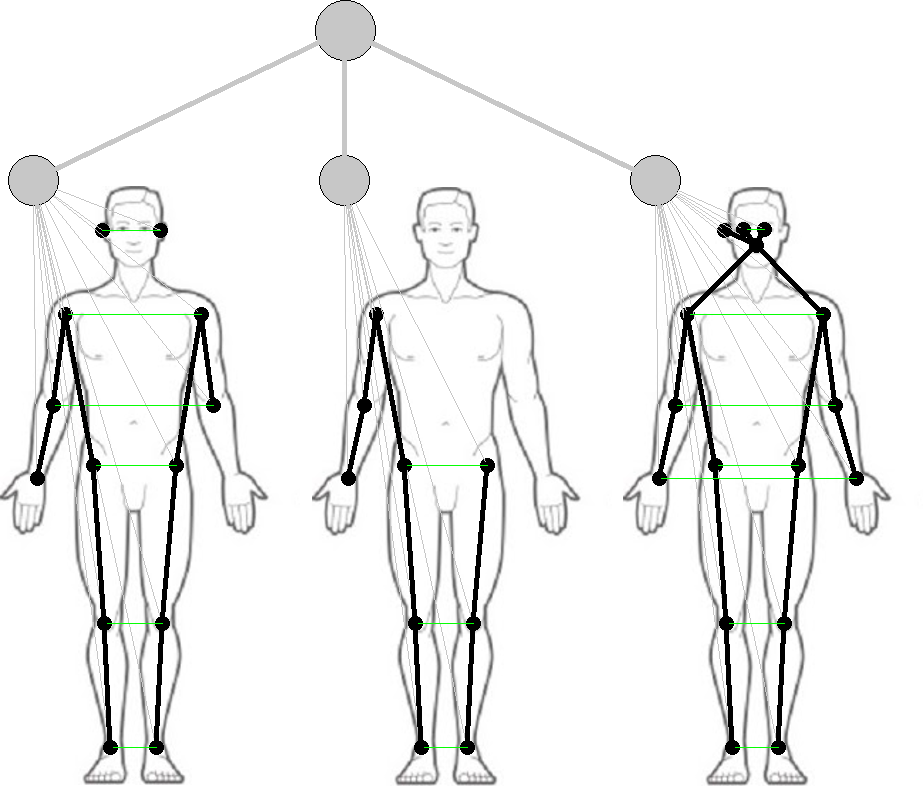}
  \label{fig:graph1}
  }
  \subfigure[Person partially detected by a single camera.]{
  \includegraphics[width=.75\columnwidth]{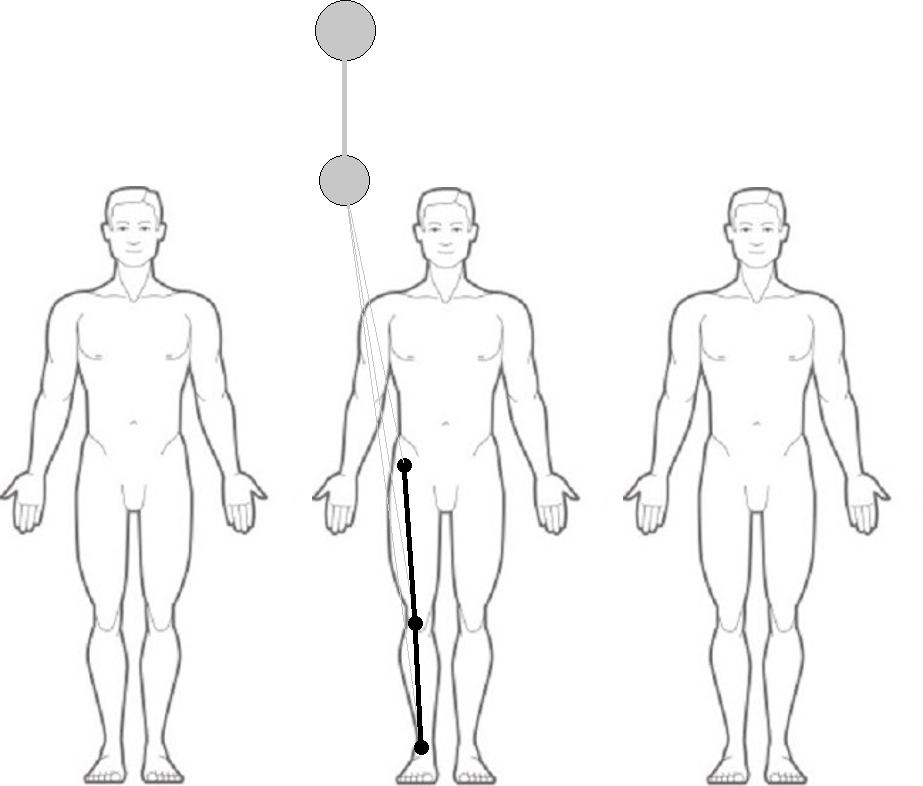}
  \label{fig:graph2}
  }
  \caption{Examples of graph representation of the information obtained by the multi-camera system.}
  \label{fig:graphs}  
\end{figure}

\textbf{GNN processing:} The models are designed to use the information obtained from each camera to estimate the position and orientation of a human even with a partial view, obtaining more accurate results when more data is available.
As can be observed in Fig.~\ref{fig:graphs}, the total number of visible joints is limited in some cases, which makes it hard to estimate the position and orientation of the person using analytical methods.
\par
GNNs adapt particularly well to structured data of varying size and missing nodes (body parts in this case)~\cite{Battaglia2018}.
Among the different GNNs variations, Graph Convolutional Networks (GCNs)~\cite{Kipf2016a} are one of the easiest to understand, and many other build on top of GCNs.
They generalise the concept of learned convolutions to graphs.
They are similar to CNNs in the sense that they learn convolutions, but instead of working on images, GNNs work on graphs.
Equation~\ref{eq:GCN} describes how the output feature vector $h_{i}^{(l+1)}$ of a node $i$ on layer $l+1$ is computed.
\begin{equation}
\label{eq:GCN}
{h}_{i}^{(l+1)} =\sigma \left ( \sum_{j\in IN(i) } \frac{1}{C_{ij}} W^{(l)} h_{j}^{(l)} \right )
\end{equation}
\par
In equation~\ref{eq:GCN}, $IN(i)$ is the set of nodes $j$ so that an edge $(j,i)$ exists in the graph, $W^{(l)}$ is the trainable weight matrix for the layer $l$, $\sigma \left ( \cdot  \right )$ is the activation function and $C_{ij}$ is as normalisation parameter.
\par
Relational Graph Convolutional Networks (RGCNs)~\cite{Schlichtkrull2018} build on top of GCNs, allowing labelled edges by using a different learnable weight matrix for each label type.
The propagation model for the feature vector of the node $i$ is shown in equation~\ref{eq:RGCN}.
\begin{equation}
\label{eq:RGCN}
h_{i}^{(l+1)}=\sigma \left ( \sum_{r\in R} \sum_{r\in N_{i}^{r}} \frac{1}{C_{i,r}} W_{r}^{(l)} h_{j}^{(l)} + W_{0}^{(l)} h_{i}^{(l)}\right )
\end{equation}
where $W_{r}^{(l)}$ and $W_{0}^{(l)}$ are the learnable matrices for \mbox{$r$-labelled} edges and self-edges, respectively.
\par

Graph Attention Networks (GATs)~\cite{Velickovic2018} introduce self-attention to GNNs.
A simplified node propagation function for the single-headed version of GAT can be seen in equation~\ref{eq:GAT}.
\begin{equation}
\label{eq:GAT}
{h}_{i}^{(l+1)} = \sigma \left (  \sum_{j\in N_{i}} \alpha_{ij}^{(l)} W^{(l)} {h}_{j}^{(l)} \right )
\end{equation}
In this case, the feature vector of node $i$ is updated from the neighbouring nodes weighted by a learnable attention parameter $\alpha$.
\par
An example of the extraction of information of body parts by a GNN can be seen in~\cite{Li2019}, where the features of the hand are encoded in a graph of different points and a GCN yields the hand gesture. 
Similarly, \cite{yan2018spatial} uses the coordinates of the human skeleton joints as the input of a GCN to recognise actions performed on videos. 
\par
In the present work, an input graph is created for each set of skeletons belonging to the same person.
First, the joints detected by each camera are used to create a separate graph corresponding to a different view of the same skeleton.
These graphs have a node representing the body and additional nodes for all the body parts available (as provided by OpenPifPaf), connecting each part not only to its kinematic parent but also to its mirrored body part.
The nodes representing the body are referred to as \textit{body nodes} in this paper.
Finally, all body nodes available (one per view) are connected to an additional node aggregating the information from the previous nodes; we call that node the \textit{superbody} node.
Fig.~\ref{fig:graphs} depicts two graphs with the body parts captured by our three-camera setup.
\par

The feature vectors of the nodes have 25 dimensions (28 if using RGBD cameras).
The feature vector for each node $i$ is built by concatenating one-hot encodings and metric information: $$h_i^{(0)} = (t_i | c_i | p_i | s_i)$$
\begin{itemize}
\item Node type one-hot encoding ($t_i$): It encodes the node type. It has a length of 19  because OpenPifPaf can detect 17 body parts and there are two additional types for the \textit{body} and \textit{superbody} nodes.
\item Camera one-hot encoding ($c_i$): This one-hot encoding represents the camera that captures the skeleton. In our experiments it has a length of 3, as that is the number of cameras used. It is zero-filled for the \textit{superbody} node.
\item Coordinates vector ($p_i$): It has a length of 2 (5 if using RGBD cameras) and stores the coordinates of each body part.
\textit{Body} and \textit{superbody} nodes have zeros in all the elements of this vector.
The image coordinates provided by OpenPifPaf are normalised so that they are within the range $[-1, 1]$.
If using 640x480 cameras, the normalisation would be:
        $$p_x' = (x-320)/320$$
        $$p_y' = (240-y)/240$$
The 3D coordinates are only provided if using RGBD cameras. They are also normalised to be in the range $[-1, 1]$, based on the size of the room. 
\item Score ($s_i$): This is a single element field that provides the certainty of the measure gathered by OpenPifPaf. It is only used for body part nodes, zero otherwise.
\end{itemize}
\par
The model is trained so that the output feature vectors are 4-dimensional and correspond to $x$, $y$, $sin(\alpha)$ and $cos(\alpha)$ for the \textit{superbody} node in the last layer.
The actual angle $\alpha$ is then reconstructed from its sine and cosine.

\section{Dataset generation}
\label{sec:dataset}
As the models are scenario-specific, they have to be trained with simulations run with the camera calibration information.
Datasets can be built using any simulator that can animate avatars, provide their ground-truth positions, and provide RGB(D) streams so that OpenPifPaf or a similar software can be used to detect people and their skeletons.
To create a proper virtual replica, the intrinsic and extrinsic parameters must be estimated.
The software released uses an Augmented Reality (AR) tag placed on the floor (so that it can be detected by multiple cameras) and guides users through the calibration process.
\par
Once the cameras are calibrated, new datasets can be easily produced generating paths for the simulated avatars in the virtual model.
Using this procedure, a big amount of data can be gathered with a limited effort.
Nevertheless, simulated data can be insufficient depending on the accuracy of the calibration because small calibration errors can lead to a significant reduction of accuracy in the estimation.
For this reason, the dataset generated from the simulations can be extended with real data, recording an actual human moving around the environment.
In our experiments, to store ground-truth information, the human was equipped with an Intel RealSense tracking camera in their chest, which provides under $1\%$ closed loop drift.
The camera pose at the start coincides with the global reference frame of the room.
Thus, the pose of the camera directly corresponds to the ground truth pose.
Combining both, simulated and real data, a final training dataset with more than 20,000 samples was created.
Specifically, 19833 samples of the dataset are simulated and 631 are real. The final dataset is provided in JSON format (available in the public repository\footnote{https://github.com/vangiel/WheresTheFellow}).
\par
It is worth mentioning that calibration is only necessary to build a replica of the real world in the simulator.
If the dataset is composed only of real data, the calibration is not necessary.

\section{EXPERIMENTAL RESULTS} \label{experiments}
To evaluate the proposed multi-camera human torso pose estimation system several experiments were conducted.
Each experiment was carried out using three different architectures: 1) a sequence of GAT layers, 2) a sequence of RGCN layers, and 3) a sequence of per-camera fully connected layers (FC) (shared across cameras) followed by concatenation and a further sequence of FC layers (MLP). For the MLP architecture, a parallel input vector of 0s and 1s was provided alongside the normal input, to indicate missing data.
\par

The three architectures were trained using the dataset described in section~\ref{sec:dataset} (\textbf{DS1}), applying different combinations of hyperparameters to select the best ones.
In addition, a second training dataset (\textbf{DS2}) including only simulated data was generated.
The three architectures were also trained using this second dataset following the same process of hyperparameter tuning.
For GNNs, different values for the number of layers, number of hidden units, attention heads (for GAT only), number of bases (for RGCN only) and activation function of each layer were tested.
These values were randomly generated to cover a wide range of combinations.
Similarly, for the MLP, we explored various depths and widths of the hidden layers.
\par
Besides the two training datasets, two additional datasets were generated from real data: one for development composed of 225 samples and another one with 283 samples for testing purposes.
Larger datasets would have been collected, but it was not possible due to the COVID-19 pandemic.

\par

Each sample of the datasets corresponds to a graph representing the view of a human from the cameras at a given instant.
Since RGBD cameras were available, for each perceived joint both 2D image coordinates and 3D positions were available.
Nevertheless, in order to make RGBD cameras optional, each architecture was trained using two versions of the data (with and without 3D information).
The first version includes the 3D and image coordinates of the body parts in the feature vectors.
The second one considers only the image coordinates so that it can be applied to multi-camera systems composed of RGB cameras.
\par

Table~\ref{tab:global_mse} shows the Mean Squared Error (MSE) of the test dataset for the best model of each architecture using the 2D-only and 3D versions of the data and the two training datasets.
As can be observed, the two GNN architectures provide better results than MLP for all the combinations of training datasets and types of features.
As expected, the use of a training dataset including only simulated data (DS2) produces a loss of accuracy in all the cases, with a more significant effect in MLP.
This fact becomes more evident with the 2D version of the data, affecting the estimation of both, orientation (see orientation MSE in table~\ref{tab:orientation_mse}) and position (see position MSE in table~\ref{tab:position_mse}).
Nevertheless, for the training dataset combining simulated and real data (DS1), the use of 2D-only features does not have a big impact on the results.


\begin{table}[ht]
\caption{Orientation MSE for the two training datasets}
\centering
\begin{tabular}{cccc}
\textbf{\hfill}	& \textbf{MLP} & \textbf{RGCN} & \textbf{GAT}\\
\hline\hline
\textbf{DS1 - 3D features} & 0.024 & 0.018 & 0.019\\
\hline
\textbf{DS1 - 2D features} & 0.026 & 0.02 & 0.021\\
\hline
\textbf{DS2 - 3D features} & 0.083 & 0.080 & 0.038\\
\hline
\textbf{DS2 - 2D features} & 0.077 & 0.058 & 0.066\\
\hline
\end{tabular}
\label{tab:orientation_mse}
\end{table}

\begin{table}[ht]
\caption{Position MSE for the two training datasets}
\centering
\begin{tabular}{cccc}
\textbf{\hfill}	& \textbf{MLP} & \textbf{RGCN} & \textbf{GAT}\\
\hline\hline
\textbf{DS1 - 3D features} & 0.0011 & 0.00079 & 0.00092\\
\hline
\textbf{DS1 - 2D features} & 0.0012 & 0.0016 & 0.0011\\
\hline
\textbf{DS2 - 3D features} & 0.0057 & 0.0021 & 0.0013\\
\hline
\textbf{DS2 - 2D features} & 0.010 & 0.0064 & 0.0049\\
\hline
\end{tabular}
\label{tab:position_mse}
\end{table}

\begin{table}[ht]
\caption{Global MSE for the two training datasets}
\centering
\begin{tabular}{cccc}
\textbf{\hfill}	& \textbf{MLP} & \textbf{RGCN} & \textbf{GAT}\\
\hline\hline
\textbf{DS1 - 3D features} & 0.010 & 0.0076 & 0.0083\\
\hline
\textbf{DS1 - 2D features} & 0.011 & 0.009 & 0.009\\
\hline
\textbf{DS2 - 3D features} & 0.037 & 0.033 & 0.016\\
\hline
\textbf{DS2 - 2D features} & 0.037 & 0.027 & 0.029\\
\hline
\end{tabular}
\label{tab:global_mse}
\end{table}

\par
To test the accuracy of the solutions, the output was compared with an analytical estimation of the human pose based on the depth data.
The position and orientation of the human in the analytical estimation are computed as follows:
\begin{itemize}
    \item Estimated position: for each camera, the position is individually estimated as the median of the positions of the joints. The final position is computed by the average of the estimations of all the cameras perceiving at least three joints of the human.
    \item Estimated orientation: for each camera, the orientation is computed from the positions of pairs of symmetric joints. Specifically, the shoulders and hips are used. The final orientation is obtained as the average of the estimations of the different cameras. If none of the cameras perceives a symmetric pair of joints, the estimation of the previous instant is maintained.
\end{itemize}
\par
\begin{figure*}
    \centering
    \includegraphics[width=0.9\textwidth]{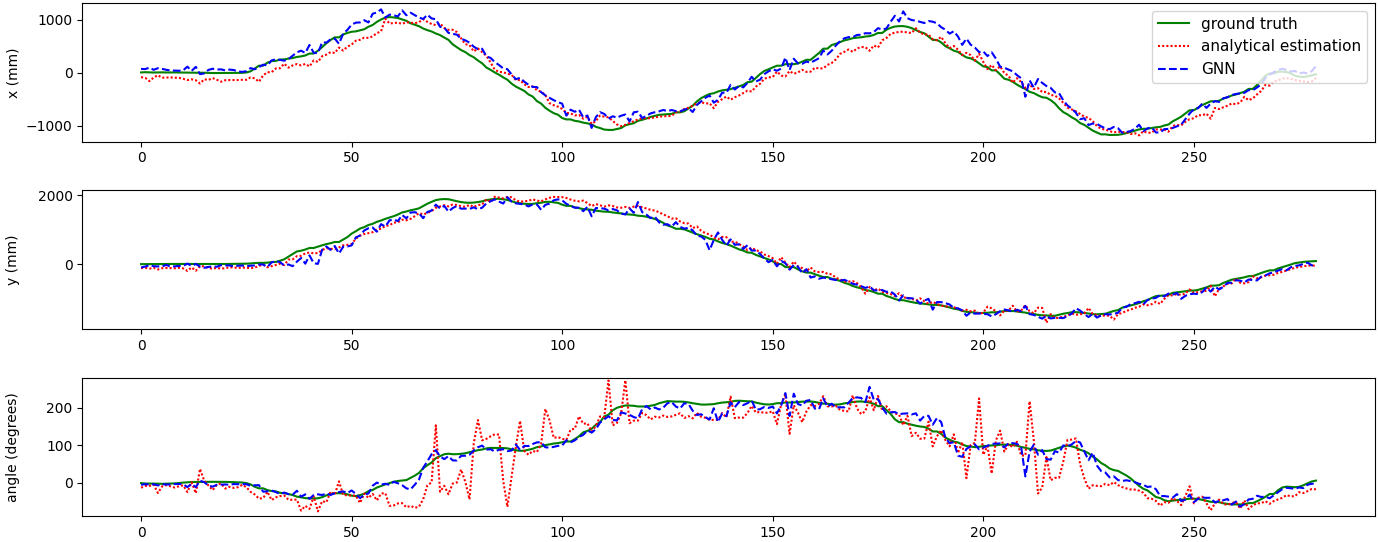}
    \caption{Comparison of the network prediction and the analytical estimation of the pose with the ground-truth for every sample of the test datasets.}
    \label{fig:test_real}
\end{figure*}
\begin{figure*}
    \centering
    \includegraphics[width=.9\textwidth]{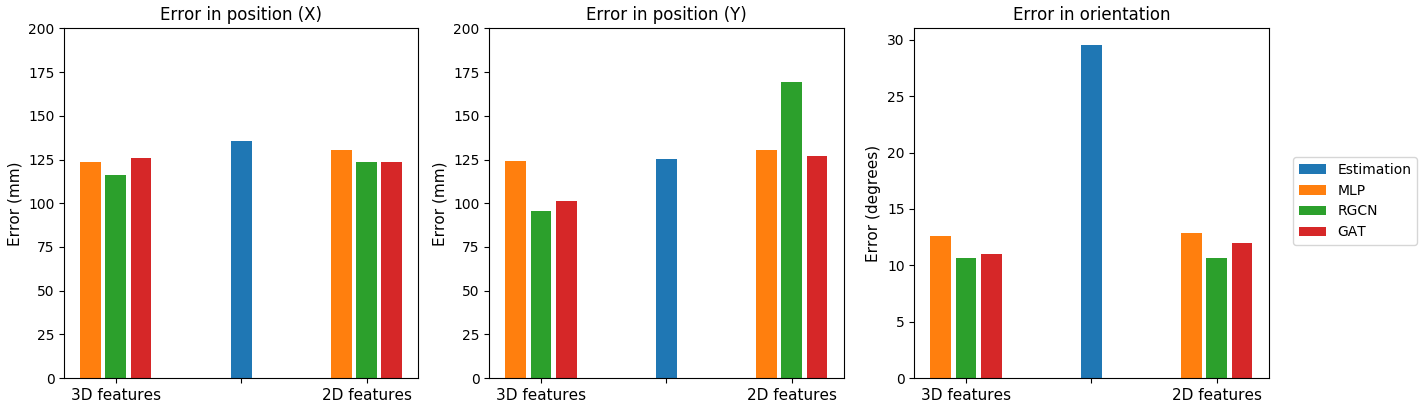}
    \caption{MAE of the position ($x$ and $y$) and orientation for the test datasets using the training dataset DS1, composed of simulated and real data.}
    \label{fig:test_all_real_DS1}
\end{figure*}

The comparison between the results obtained from the analytical and the learnt estimators shows how the learning-based solutions outperform the analytical method, more notably if they have access to the depth channel of the images too, especially for the orientation.
This can be observed in figure~\ref{fig:test_real}, which depicts the estimation of the position and the orientation for every sample of the test dataset using the analytical method (red dotted line) and the RGCN-based architecture trained using 3D data with the dataset DS1 (blue dashed line).
As can be seen, the difference with the ground-truth (green solid line) is smaller in the estimation obtained by the GNN, which is especially remarkable in the angle prediction.
This observation can be extended to the remaining architectures trained with DS1 (figure~\ref{fig:test_all_real_DS1}).
In fact, the mean absolute error (MAE) of the best GNN architecture considerably outperforms the analytical method, providing a mean absolute angle error below $10^{\circ}$.
\par
\begin{figure*}
    \centering
    \includegraphics[width=.9\textwidth]{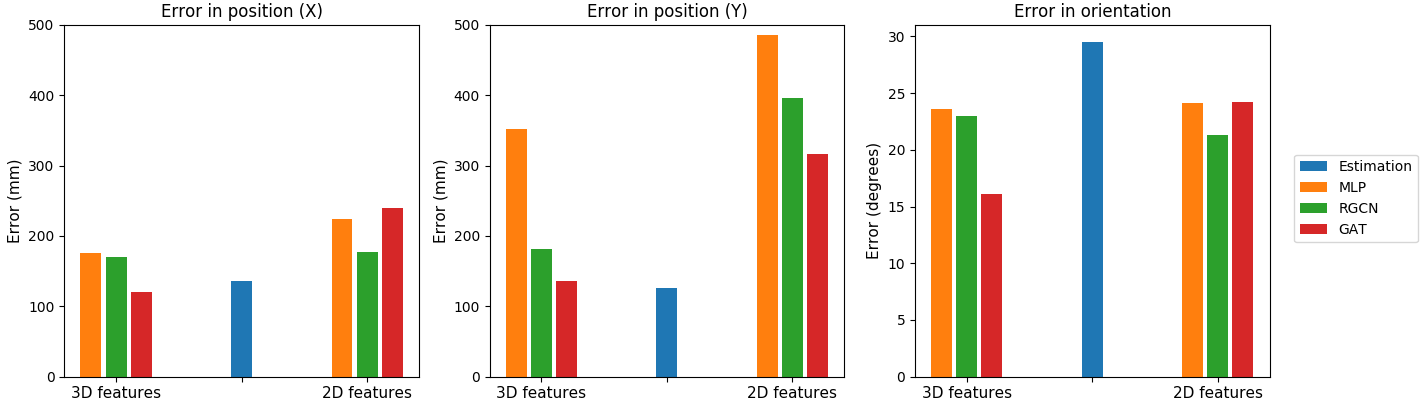}
    \caption{MAE of the position ($x$ and $y$) and orientation for the test datasets using the training dataset DS2 composed entirely of simulated data.}
    \label{fig:test_all_real_DS2}
\end{figure*}

The exclusive use of simulated data for training produces a deterioration of the results which can be seen in figure~\ref{fig:test_all_real_DS2}.
However, the use of 3D information in the data (3D version of DS2) still outperforms the analytical estimation for one of the GNN architectures in position and orientation.

\section{CONCLUSIONS}
Our human pose estimation system has been designed for spaces that are covered by multiple cameras. It is assumed that people are visible by at least one of the cameras, although some parts of their bodies can be occluded or outside of the field of vision of some of the cameras.
\par
In comparison to other works, our approach outperforms~\cite{Shinmura2015} and it is -under good conditions- outperformed by~\cite{Lewandowski2020}.
Although~\cite{Lewandowski2020} reports better results, it requires RGBD cameras, which are an order of magnitude more expensive than low-resolution RGB cameras.
Additionally, as reported in~\cite{Lewandowski2020} their dataset does not consider occlusions or partial views, so their results will likely deteriorate in real-life conditions.
\par
A mean absolute error of $125 mm$ in the pose's coordinates seems reasonable for most human-robot interaction tasks such as human-aware navigation, considering: \textbf{a)}~the size of the average human \textit{i.e.}, the percentile 50 of the forearm-forearm breadth of an adult is about $492 mm$ (female) and $579 mm$ (male)~\cite{gordon20142012} and \textbf{b)}~proxemics studies have reported personal spaces to approximate to a circle of about $1200 mm$~\cite{Pacchierotti2005}.
\par
Given the accuracy achieved using regular RGB images and the limited improvements obtained from the use of the depth channel, regular low-end webcams seem sufficient for most HRI scenarios.
\par
Camera calibration can be avoided if the dataset is exclusively generated from real scenarios. Nevertheless, the use of a realistic simulator to generate most of the data used for training drastically reduces the time and resources needed to obtain a valid solution to the pose estimation problem. A fraction of data from the real set up seems necessary to account for some calibration and modelling errors but future works aim at reducing this ratio even more.



\bibliographystyle{IEEEtran}
\bibliography{bib.bib}

\begin{thebibliography}{10}
\providecommand{\url}[1]{#1}
\csname url@rmstyle\endcsname
\providecommand{\newblock}{\relax}
\providecommand{\bibinfo}[2]{#2}
\providecommand\BIBentrySTDinterwordspacing{\spaceskip=0pt\relax}
\providecommand\BIBentryALTinterwordstretchfactor{4}
\providecommand\BIBentryALTinterwordspacing{\spaceskip=\fontdimen2\font plus
\BIBentryALTinterwordstretchfactor\fontdimen3\font minus
  \fontdimen4\font\relax}
\providecommand\BIBforeignlanguage[2]{{%
\expandafter\ifx\csname l@#1\endcsname\relax
\typeout{** WARNING: IEEEtran.bst: No hyphenation pattern has been}%
\typeout{** loaded for the language `#1'. Using the pattern for}%
\typeout{** the default language instead.}%
\else
\language=\csname l@#1\endcsname
\fi
#2}}

\bibitem{Glonek2017}
G.~Glonek and A.~Wojciechowski, ``{Hybrid orientation based human limbs motion
  tracking method},'' \emph{Sensors}, vol.~17, no.~12, 2017.

\bibitem{Thang2019}
{D. N. Thang et al.}, ``{Deep Learning-based Multiple Objects Detection and
  Tracking System for Socially Aware Mobile Robot Navigation Framework},''
  \emph{NICS 2018 - Proceedings of 2018 5th NAFOSTED Conference on Information
  and Computer Science}, pp. 436--441, 2019.

\bibitem{manso2019graph}
L.~J. Manso, R.~R. Jorvekar, D.~R. Faria, P.~Bustos, and P.~Bachiller, ``{Graph
  Neural Networks for Human-aware Social Navigation},'' \emph{arXiv preprint
  arXiv:1909.09003}, 2019.

\bibitem{Mateus2019}
\BIBentryALTinterwordspacing
A.~Mateus, D.~Ribeiro, P.~Miraldo, and J.~C. Nascimento, ``{Efficient and
  robust Pedestrian Detection using Deep Learning for Human-Aware
  Navigation},'' \emph{Robotics and Autonomous Systems}, vol. 113, pp. 23--37,
  2019. [Online]. Available: \url{https://doi.org/10.1016/j.robot.2018.12.007}
\BIBentrySTDinterwordspacing

\bibitem{Shinmura2015}
F.~Shinmura, D.~Deguchi, I.~Ide, H.~Murase, and H.~Fujiyoshi, ``{Estimation of
  human orientation using coaxial RGB-depth images},'' \emph{VISAPP 2015 - 10th
  International Conference on Computer Vision Theory and Applications;
  VISIGRAPP, Proceedings}, vol.~2, pp. 113--120, 2015.

\bibitem{Wahyu2019}
R.~Wahyu and A.~Saputra, ``{Human Body ' s Orientation Estimation Based On
  Depth Image},'' \emph{2019 International Electronics Symposium (IES)}, pp.
  266--271, 2019.

\bibitem{Lewandowski2020}
B.~Lewandowski, D.~Seichter, T.~Wengefeld, L.~Pfennig, H.~Drumm, and H.-M.
  Gross, ``{Deep orientation: Fast and Robust Upper Body orientation Estimation
  for Mobile Robotic Applications},'' vol.~2, no.~03, pp. 441--448, 2020.

\bibitem{Choi2016}
\BIBentryALTinterwordspacing
J.~Choi, B.-J. Lee, and B.-T. Zhang, ``{Human Body Orientation Estimation using
  Convolutional Neural Network},'' 2016. [Online]. Available:
  \url{http://arxiv.org/abs/1609.01984}
\BIBentrySTDinterwordspacing

\bibitem{Pfinder}
C.~R. {Wren}, A.~{Azarbayejani}, T.~{Darrell}, and A.~P. {Pentland}, ``Pfinder:
  real-time tracking of the human body,'' \emph{IEEE Transactions on Pattern
  Analysis and Machine Intelligence}, vol.~19, no.~7, pp. 780--785, July 1997.

\bibitem{Papageorgiou2000}
P.~Constantine and P.~Tomaso, ``{A Trainable System for Object Detection},''
  \emph{International Journal of Computer Vision}, vol.~38, no.~1, pp. 15--33,
  2000.

\bibitem{shotton2011real}
J.~Shotton, A.~Fitzgibbon, M.~Cook, T.~Sharp, M.~Finocchio, R.~Moore,
  A.~Kipman, and A.~Blake, ``Real-time human pose recognition in parts from
  single depth images,'' in \emph{CVPR 2011}.\hskip 1em plus 0.5em minus
  0.4em\relax Ieee, 2011, pp. 1297--1304.

\bibitem{liu2015detecting}
J.~Liu, Y.~Liu, G.~Zhang, P.~Zhu, and Y.~Q. Chen, ``{Detecting and tracking
  people in real time with RGB-D camera},'' \emph{Pattern Recognition Letters},
  vol.~53, pp. 16--23, 2015.

\bibitem{calderita2013model}
L.~V. Calderita, J.~P. Bandera, P.~Bustos, and A.~Skiadopoulos, ``{Model-based
  reinforcement of Kinect depth data for human motion capture applications},''
  \emph{Sensors}, vol.~13, no.~7, pp. 8835--8855, 2013.

\bibitem{cao2018openpose}
Z.~Cao, G.~Hidalgo, T.~Simon, S.-E. Wei, and Y.~Sheikh, ``Open{P}ose: realtime
  multi-person 2{D} pose estimation using {P}art {A}ffinity {F}ields,'' in
  \emph{arXiv preprint arXiv:1812.08008}, 2018.

\bibitem{kreiss2019pifpaf}
S.~Kreiss, L.~Bertoni, and A.~Alahi, ``Pifpaf: Composite fields for human pose
  estimation,'' in \emph{The IEEE Conference on Computer Vision and Pattern
  Recognition (CVPR)}, June 2019.

\bibitem{rohmer2013coppeliasim}
E.~Rohmer, S.~P. Singh, and M.~Freese, ``{Coppeliasim: A versatile and scalable
  robot simulation framework},'' in \emph{Proc. Int. Conf. on Intelligent
  Robots and Systems}, 2013, pp. 1321--1326.

\bibitem{kailath1967divergence}
T.~Kailath, ``{The divergence and Bhattacharyya distance measures in signal
  selection},'' \emph{IEEE transactions on communication technology}, vol.~15,
  no.~1, pp. 52--60, 1967.

\bibitem{Battaglia2018}
\BIBentryALTinterwordspacing
{P. W. Battaglia et al.}, ``{Relational inductive biases, deep learning, and
  graph networks},'' pp. 1--40, 2018. [Online]. Available:
  \url{http://arxiv.org/abs/1806.01261}
\BIBentrySTDinterwordspacing

\bibitem{Kipf2016a}
\BIBentryALTinterwordspacing
T.~N. Kipf and M.~Welling, ``{Semi-Supervised Classification with Graph
  Convolutional Networks},'' pp. 1--14, 2016. [Online]. Available:
  \url{http://arxiv.org/abs/1609.02907}
\BIBentrySTDinterwordspacing

\bibitem{Schlichtkrull2018}
M.~Schlichtkrull, T.~N. Kipf, P.~Bloem, R.~van~den Berg, I.~Titov, and
  M.~Welling, ``{Modeling Relational Data with Graph Convolutional Networks},''
  \emph{Lecture Notes in Computer Science}, vol. 10843 LNCS, no.~1, pp.
  593--607, 2018.

\bibitem{Velickovic2018}
\BIBentryALTinterwordspacing
P.~Veli{\v{c}}kovi{\'{c}}, G.~Cucurull, A.~Casanova, A.~Romero, P.~Li{\`{o}},
  and Y.~Bengio, ``{Graph Attention Networks},'' in \emph{Proceedings of the
  International Conference on Learning Representations 2018}, no. 2015, 2018,
  pp. 1--11. [Online]. Available: \url{http://arxiv.org/abs/1710.10903}
\BIBentrySTDinterwordspacing

\bibitem{Li2019}
Y.~Li, Z.~He, X.~Ye, Z.~He, and K.~Han, ``{Spatial temporal graph convolutional
  networks for skeleton-based dynamic hand gesture recognition},''
  \emph{Eurasip Journal on Image and Video Processing}, vol. 2019, no.~1, 2019.

\bibitem{yan2018spatial}
S.~Yan, Y.~Xiong, and D.~Lin, ``Spatial temporal graph convolutional networks
  for skeleton-based action recognition,'' in \emph{Thirty-second AAAI
  conference on artificial intelligence}, 2018.

\bibitem{gordon20142012}
C.~C. Gordon, C.~L. Blackwell, B.~Bradtmiller, J.~L. Parham, P.~Barrientos,
  S.~P. Paquette, B.~D. Corner, J.~M. Carson, J.~C. Venezia, B.~M. Rockwell,
  \emph{et~al.}, ``2012 anthropometric survey of us army personnel: Methods and
  summary statistics,'' Army Natick Soldier Research Development and
  Engineering Center MA, Tech. Rep., 2014.

\bibitem{Pacchierotti2005}
E.~Pacchierotti, H.~I. Christensen, and P.~Jensfelt, ``{Human-robot embodied
  interaction in hallway settings: A pilot user study},'' in \emph{IEEE
  International Workshop on Robot and Human Interactive Communication}, vol.
  2005.\hskip 1em plus 0.5em minus 0.4em\relax IEEE, 2005, pp. 164--171.

\end{thebibliography}

\end{document}